\documentclass[]{fairmeta}
\usepackage{tikz}
\usetikzlibrary{positioning,arrows.meta,calc,fit,backgrounds}
\usepackage{amsmath}
\usepackage{amssymb}
\usepackage{wrapfig}

\title{Look Before You Launch: \underline{L}earned Throughput Pr\underline{e}diction for H\underline{a}rdware Selection in a \underline{P}roduction Training Fleet}
\hypersetup{pdftitle={Look Before You Launch: Learned Throughput Prediction for Hardware Selection in a Production Training Fleet}}

\author[1]{Hossein Askari}
\author[*]{Wei Sun}
\author[1]{Chenguang Zhu}
\author[1]{Bin Zan}
\author[*]{Sreen Tallam}
\author[1]{Keyu Man}

\affiliation[1]{Meta}
\affiliation[*]{Work done at Meta}

\abstract{
In mixed-vendor accelerator fleets, many training jobs can run on several
hardware types, and the best choice depends on each job's
throughput on each type.
Routing a job needs a throughput estimate for every candidate type before the
job runs.
We present LEAP, a machine learning performance prediction service that
predicts each training job's throughput on every candidate hardware type at
submission time and ranks the candidates by predicted throughput per unit cost. It routes tens of thousands of
training job attempts a day across hundreds of different model types and more than 20
hardware types in a production fleet. LEAP predicts throughput from submission-time
training configuration, including each job's prior throughput.
LEAP reaches an accuracy of 77\% over 30 nightly retrains, higher
than the lookup table it replaced and at least as high as every alternative
model we tested, including fine-tuned language models. Over six months, the share of eligible training GPU-hours that received a ranking by LEAP
grew from 58\% to 95\%, and we estimate that it saves about 3\% of training
computational cost. In this paper, we identify the drivers that actually move accuracy and those that do not. We offer these as lessons for teams building performance predictors at fleet scale.
}

\date{\today}
\correspondence{Hossein Askari at \email{haskari@meta.com}}

\begin{document}

\maketitle

\section{Introduction}
\label{sec:intro}

Large training fleets increasingly mix accelerators from several vendors
and generations, and many training jobs can run on more than one of them. Because candidate hardware types vary significantly in both computational cost and job throughput (QPS), hardware choice directly dictates job computational cost. However, because jobs placed on sub-optimal hardware still complete successfully without reporting errors, efficiency losses often go unnoticed. Forecasting pre-execution throughput across candidate hardware introduces a formidable fleet-scheduling challenge, particularly given that hardware routing governs roughly half of the fleet’s half-year GPU-hours. 

\begin{wrapfigure}{r}{0.36\textwidth}
\centering
\vspace{-1.2em}
\includegraphics[width=0.36\textwidth]{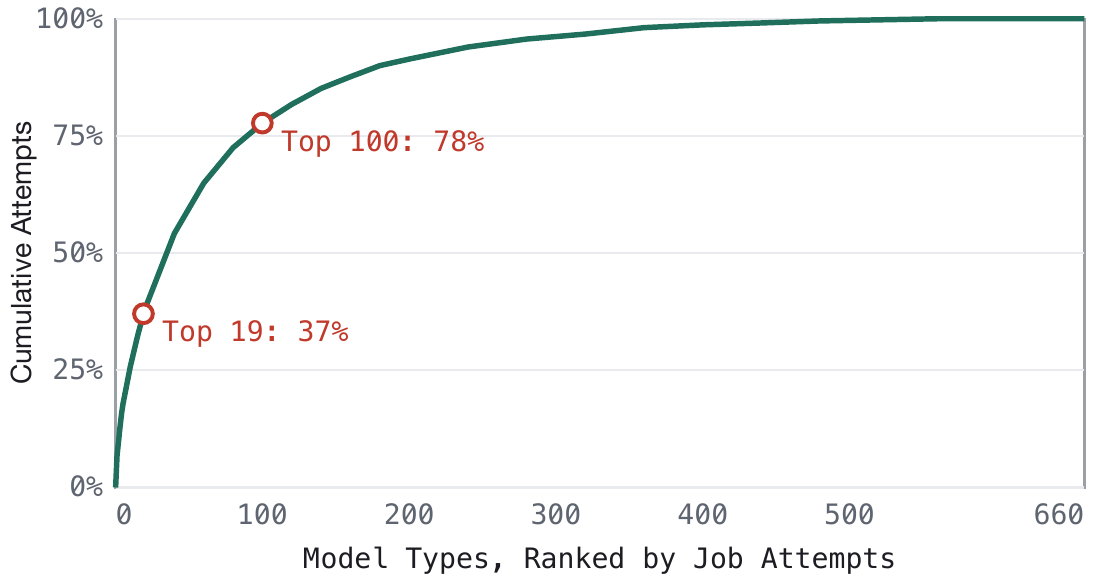}
\caption{Routing must serve a long tail of model types. Cumulative share of
one day's job attempts over model types ranked by volume.}
\label{fig:longtail}
\vspace{-1em}
\end{wrapfigure}
The legacy in-house solution, relied on a zero-ML lookup table keyed on (\texttt{model\_type}, \texttt{num\_trainers}, \texttt{batch\_size}, \texttt{hardware\_type}). It suffered from low coverage and rigid heuristics. To solve this, LEAP replaces it with a learned prediction service, which dynamically constructs per-job preference lists and routes execution via the cluster scheduler to maximize performance per unit of infrastructure computational cost.

Routing a fungible job, one that can run on any of several hardware types, requires an estimate of the training throughput (samples per second) it would achieve on each candidate type. The scheduler needs these estimates at submission time, before the current attempt has executed. Past execution attempts yield sparse historical signals, covering only previously encountered hardware configurations while offering zero coverage for novel workloads or unvisited accelerator types. The workload is also heavily long-tailed. On average, in a typical day, the fleet executes 25,000 job attempts drawn from approximately 660 production model types. On the day shown in Figure~\ref{fig:longtail}, the 560 least frequent model types together account for 22\% of attempts. A routing policy built only for the most common model types would therefore leave a substantial share of the workload unserved.

\begin{figure}[t]
\centering
\includegraphics[width=\textwidth]{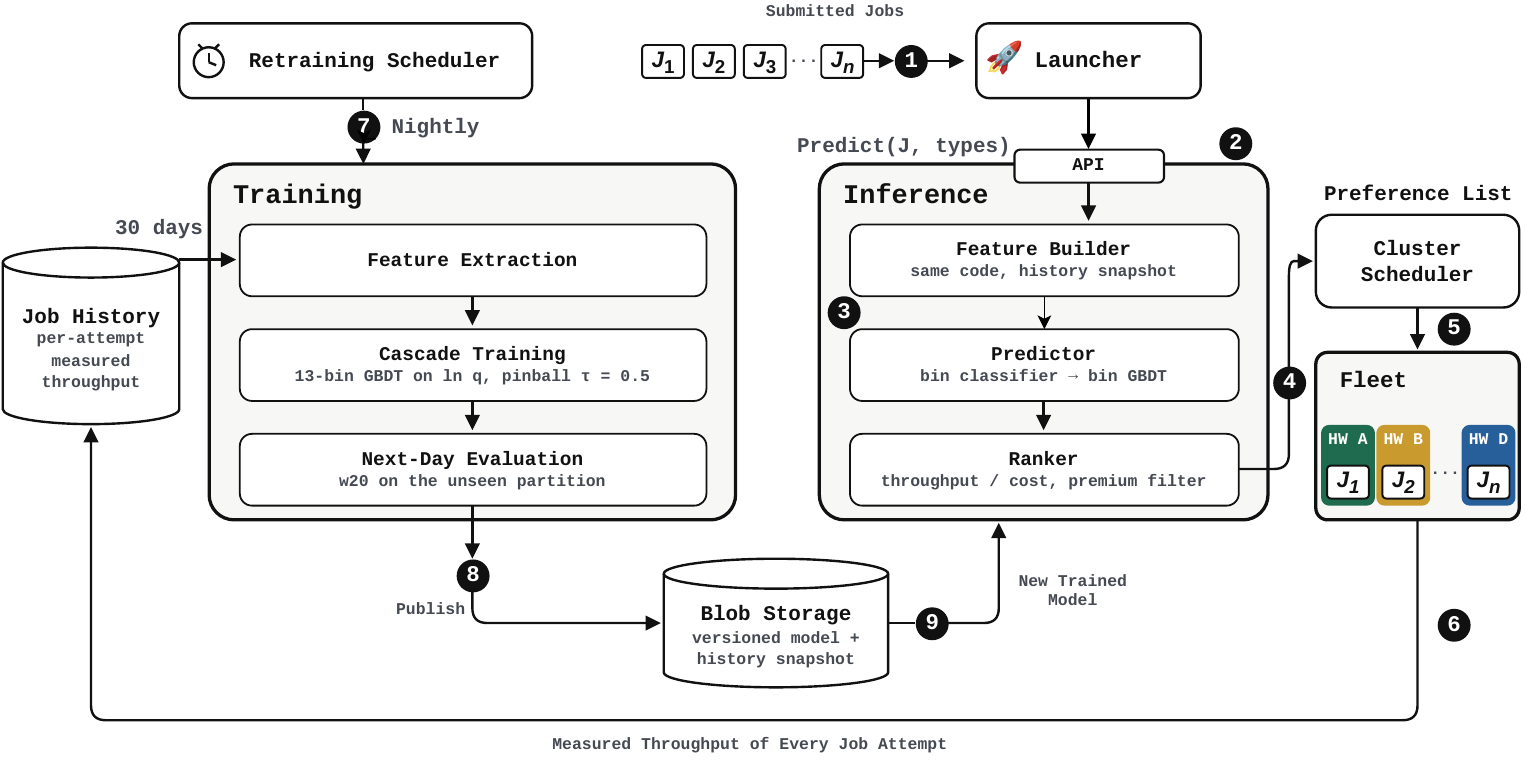}
  \caption{LEAP architecture. Serving: (1) jobs are submitted to the job
  launcher, which (2) calls the inference API with the job and its candidate
  hardware types. (3) Inference builds the job's features with the same code as
  training, taking its prior throughput from the history snapshot, scores every
  candidate with the bin classifier and that bin's gradient-boosted model, and
  ranks candidates by predicted throughput per unit cost. (4) The preference list
  goes to the cluster scheduler, which (5) places the job on the fleet. Training:
  (6) each completed job attempt records its measured throughput in the job
  history; (7) a nightly trigger retrains the 13-bin cascade on the last 30 days
  and evaluates it on the next, unseen day; (8) the run publishes a versioned
  model and history snapshot to blob storage; and (9) inference loads the latest
  version through a stable pointer.}
\label{fig:system}
\end{figure}

\begin{figure}[t]
\centering
\includegraphics[width=\textwidth]{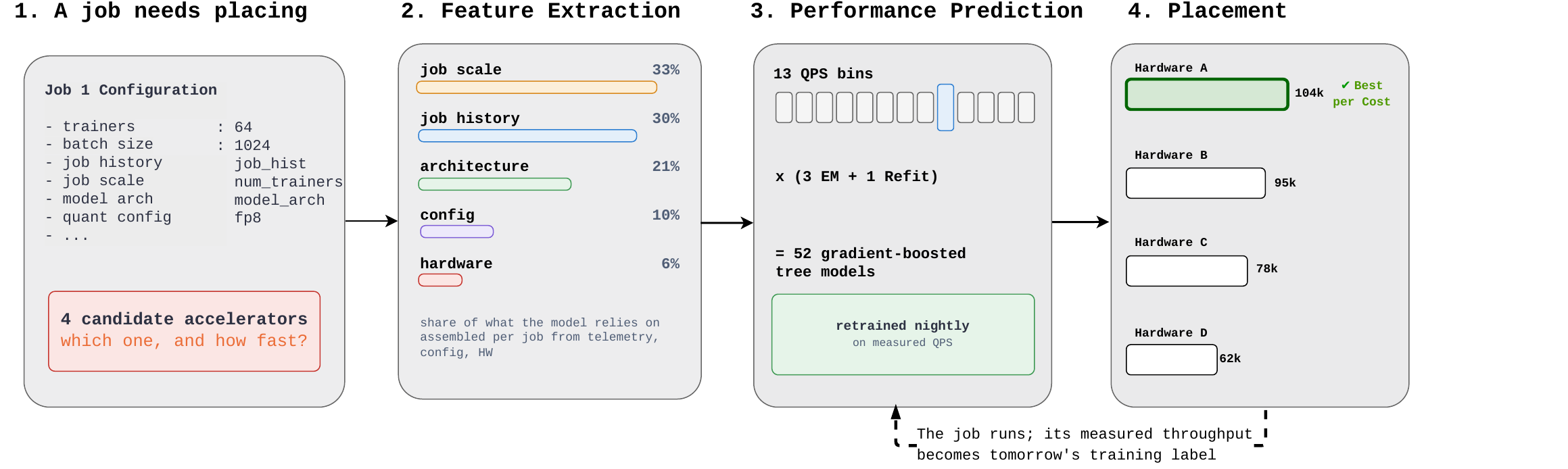}
\caption{LEAP end to end. (1)~A fungible job arrives with several
candidate hardware types. (2)~LEAP builds features from the job's
configuration, its prior throughput, and each type's specifications; bars give
each feature group's share of importance. (3)~A 13-bin cascade of
gradient-boosted models, retrained nightly, predicts the job's throughput on
every candidate. (4)~The ranking stage orders the candidates for the cluster
scheduler. The job's measured throughput becomes training data for the next
night's models.}
\label{fig:LEAP_system}
\end{figure}

Existing approaches do not provide this estimate at submission time without additional computational cost. Performance models such as Habitat and DNNPerf need either a profiling run of the job on available hardware or its computation graph~\citep{yu2021habitat, gao2023dnnperf}. Heterogeneity-aware schedulers such as Gavel and Sia either assume per-type throughput is known in advance or estimate it from measurements taken while jobs run~\citep{narayanan2020gavel, jayaram2023sia}. Our initial routing method was a lookup table keyed on model type, number of trainers (hosts per job), batch size, and hardware type. This key does not determine throughput. For 56\% of job attempts, the 10th-to-90th percentile throughput range under the attempt's key differs by more than a factor of 1.5, so no single prediction can fall within $\pm$20\% across that range. Coverage was also limited, the table contained entries for only 19 model types, and on a held-out day it produced an estimate for 13\% of job attempts.

LEAP replaces the lookup table with a learned model. Instead of matching a
job to a stored entry, it predicts the job's throughput on each candidate
hardware type from features available at submission such as the job's training
configuration, the hardware type, and the throughput
the same job achieved in earlier runs. LEAP then ranks the candidates by
predicted throughput per unit cost and passes the ranking to the cluster
scheduler. The model is retrained every night on the measured throughput of
jobs that have finished.

Figure~\ref{fig:system} shows how LEAP operates. When a job
is submitted, the job launcher calls LEAP's inference API with the job's
configuration and its candidate hardware types. LEAP builds the job's
features with the same code used in training, adds the job's prior
throughput from a history snapshot stored with the model, predicts the
job's throughput on every candidate, and ranks the candidates by predicted
throughput per unit cost. The cluster scheduler receives this ranking as a
preference list and places the job on the fleet. After the job runs, its
measured throughput is recorded in the job history. A nightly pipeline
retrains the model on the last 30 days of this history, evaluates it on the
following day, and publishes a new versioned model and history snapshot,
which the inference API loads through a stable pointer. Each placed job
therefore becomes training data for the next day's model.

Over the first half of 2026, the share of eligible training GPU-hours
that received a ranking from LEAP grew from 58\% to 95\% (77 million GPU-hours), and we
estimate that it reduces training computational cost by about 3\%. We measure accuracy as
the share of jobs whose predicted throughput is within $\pm 20\%$ of the
measured value. The lookup table returned an estimate for only 13\% of
jobs, while LEAP returns one for every job it receives. On the jobs the
table covers, LEAP's accuracy is 85\%, compared with 53\% for the table
(Section~\ref{sec:baseline}). We also compared LEAP's gradient-boosted
model with other model families on the same data. LightGBM, TabPFN,
fine-tuned language models such as Gemma~3~\citep{gemma2025gemma3} and Llama~4~\citep{meta2025llama4}, and language-model
encoders were all less accurate. A graph neural network built on the job's
computation graph matched LEAP's accuracy but needs a GPU and a large graph
cache at serving time (Sections~\ref{sec:levers} and~\ref{sec:llm}). In
the following sections, we describe each component of LEAP in detail.
\section{System Design}
\label{sec:design}
Deploying a learned predictor in place of the lookup table showed that
throughput prediction is only one stage of a longer pipeline between job
submission and placement, which we call the \emph{routing chain}. The job
launcher first determines which hardware types can host the job. LEAP then
predicts the job's throughput on each eligible type and ranks the types by predicted throughput per unit cost. Finally, the cluster scheduler treats the ranking as a preference,
not a binding constraint, and places the job according to available
capacity. LEAP implements the prediction and ranking stages, and the
surrounding stages belong to other systems.

Within these boundaries, LEAP consists of four components
(Figures~\ref{fig:system} and~\ref{fig:LEAP_system}). \emph{Feature
extraction} (\ref{sec:features}) converts a job's launch configuration into
one feature row per candidate hardware type, with a single implementation
shared by training and serving. \emph{Performance prediction}
(\ref{sec:model}) estimates throughput with a binned cascade of
gradient-boosted trees in log space, because throughput spans several
orders of magnitude and a single regressor fits the extremes poorly.
\emph{Placement} (\ref{sec:serving}) runs inside the launcher, divides each
prediction by the computational cost of the hardware type, removes premium types unless
the job is predicted to use them efficiently, and returns the ranked
preference list. Runtime switches enable routing one model type at a time.
\emph{Nightly retraining} (\ref{sec:training}) refits the models on the
measured throughput of completed jobs and publishes them as versioned
artifacts behind a stable pointer, so launchers pick up a new model without
a code change. Because a job has not yet run when it is submitted, every
serving-time input must come from its static configuration, the hardware
specification table, or data shipped inside the model artifact.

\subsection{Feature Extraction}
\label{sec:features}

Training data come from a daily-partitioned feature table. On average, each partition
contains 25,000 job attempts spanning more than 660 model types
and more than 20 hardware types. Each row joins three sources: job-attempt
records (launch configuration and measured outcomes), cluster allocation logs
(provisioned resources), and a static hardware specification table.  A single extraction module, used unchanged by the training pipeline and the serving path, computes the features, and measured throughput serves as the label.

Hardware types are represented by twelve numeric specifications (e.g., peak
compute throughput, memory capacity and bandwidth, and interconnect bandwidth)
plus a vendor identifier. A new hardware type therefore enters the model as a new point in a familiar feature space, although in practice the model needs production runs on it before its predictions are reliable (Section ~\ref{sec:unseen-hw}).

\begin{figure}[t]
\centering
\begin{minipage}[t]{0.485\textwidth}
\centering
\includegraphics[width=\linewidth]{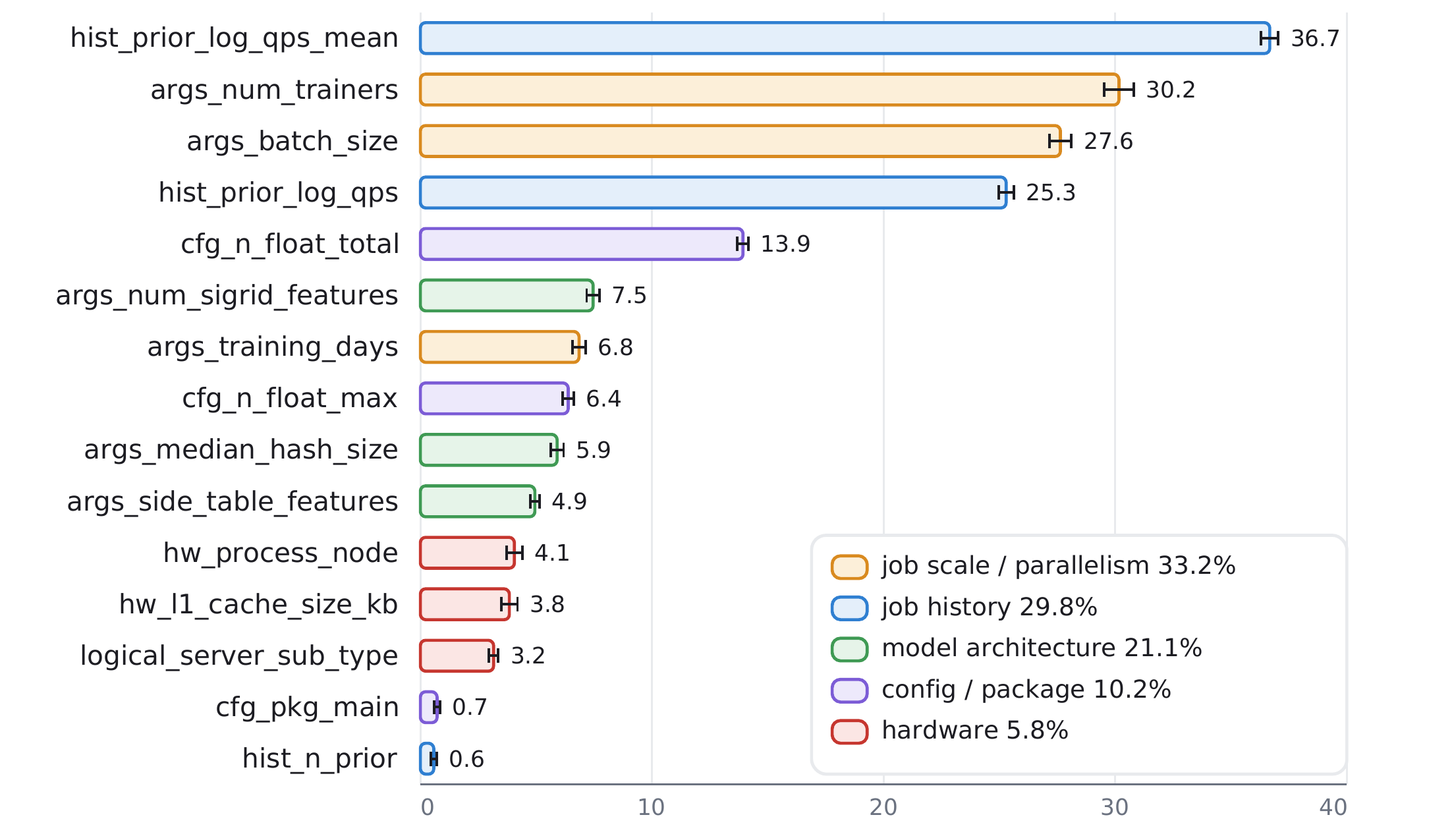}
\caption{Permutation importance, measured as the drop in accuracy (share of
jobs within $\pm$20\% of measured throughput) when a single column is
shuffled, averaged over 10 shuffles on a held-out subset of one day. By group, job scale and job history account for
63\% of total importance and hardware specifications for 5.8\%; 18 of the 73
columns have no measurable effect.}
\label{fig:featimp}
\end{minipage}\hfill
\begin{minipage}[t]{0.485\textwidth}
\centering
\includegraphics[width=\linewidth]{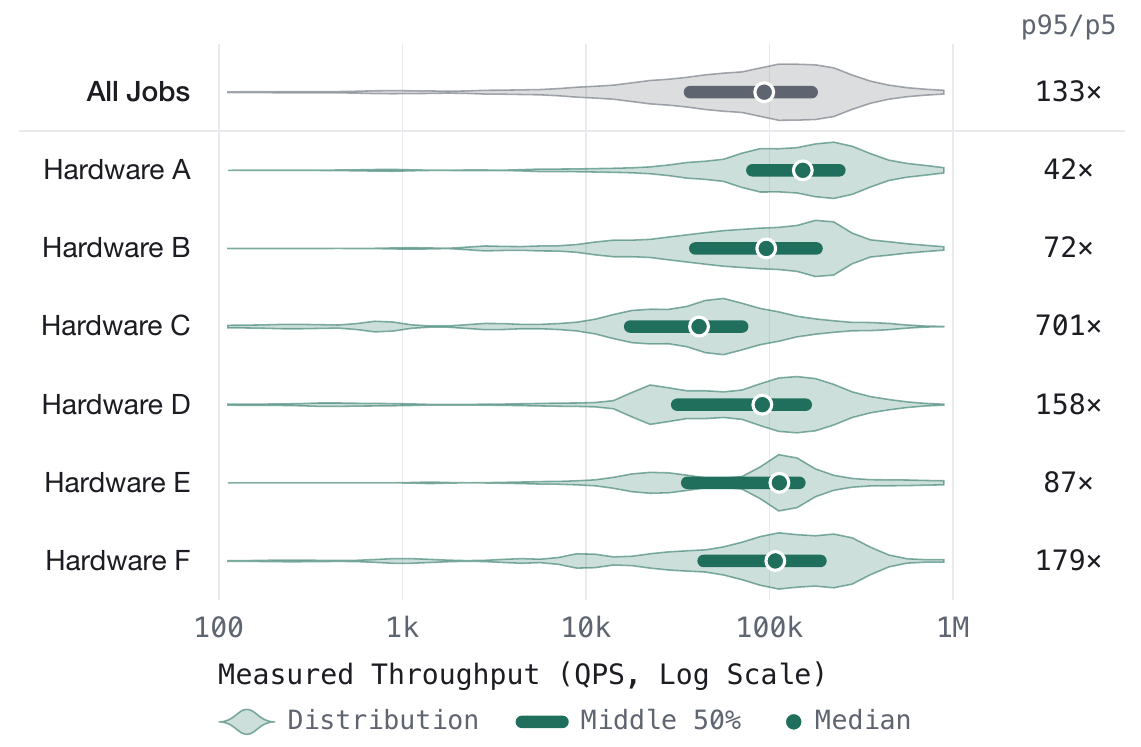}
\caption{Measured throughput (log scale) of models per hardware
type. Bars mark the interquartile range and dots the median. The right
column gives the p95/p5 ratio.}
\label{fig:qps-range}
\end{minipage}
\end{figure}

Figure~\ref{fig:featimp} shows permutation importance for the most
influential features, measured as the drop in accuracy when a single feature
is shuffled on a held-out day and the trained model is re-scored. The mean of
the job's prior log throughput has the largest effect (36.7 points), followed
by trainer count (30.2) and batch size (27.6). The most recent prior log
throughput ranks fourth (25.3). Aggregated by category, job scale accounts
for 33.2\% of total importance and job history for 29.8\%. Hardware
specifications account for 5.8\%, which is low given that hardware type is
what the prediction is used to select.

Permutation importance should be interpreted with care. It reflects how much
the trained model relies on a feature, not how much accuracy would be lost
if the feature were removed, because a model retrained without history
features can recover part of that signal from correlated inputs such as
model architecture and job scale. Correlated features also divide credit
among themselves. The hardware specifications vary together across a small number of hardware types, so shuffling any one of them leaves most of the hardware information intact, and the 5.8 \% figure
  likely understates the contribution of hardware as a whole.

\subsection{Prediction Model}
\label{sec:model}

Throughput is heavily skewed. On one partition of 22,495 rows, measured
throughput spans seven orders of magnitude, from single digits to just over
$10^{6}$, yet 91\% of rows fall within two of those decades.
Figure~\ref{fig:qps-range} shows that this spread is not an artifact of
pooling hardware types. On every hardware type, the ratio of 95th- to
5th-percentile throughput ranges from 42$\times$ to 701$\times$
(133$\times$ across all jobs). Hardware type shifts the distribution, with
median throughput on Hardware~D roughly $3\times$ lower than on Hardware~B,
but it does not remove the spread. A regressor fit on the raw scale is
dominated by the largest values, and absolute error is not a meaningful
objective across such a range.

LEAP therefore predicts the log target $y = \ln q$, where $q$ is measured
throughput, and trains each regressor $f$ under the pinball loss at the
median. With residual $u = y - f(x)$,
\begin{equation}
\ell_{\tau}(u) = u\,\big(\tau - \mathbf{1}[u < 0]\big), \qquad \tau = 0.5 .
\label{eq:pinball}
\end{equation}
At $\tau = 0.5$ the loss equals $\tfrac{1}{2}\lvert u \rvert$, so $f$
estimates the conditional median of $y$. Because the median is preserved
under monotone transformations, $\exp f(x)$ estimates the conditional median
of $q$, which lies below the conditional mean for a right-skewed
distribution.

\begin{figure}[t]
\centering
\resizebox{0.9\textwidth}{!}{%
\begin{tikzpicture}[
  font=\scriptsize,
  >={Stealth[length=4pt]},
  arr/.style={->, thick},
  box/.style={draw, rounded corners=2pt, align=center, inner sep=3pt, fill=white},
  tn/.style={circle, fill=black!65, inner sep=0pt, minimum size=2.2pt},
  tree/.pic={
    \node[tn] at (0,0) {};
    \node[tn] at (-0.13,-0.14) {}; \node[tn] at (0.13,-0.14) {};
    \node[tn] at (-0.2,-0.28) {}; \node[tn] at (-0.06,-0.28) {};
    \node[tn] at (0.06,-0.28) {}; \node[tn] at (0.2,-0.28) {};
    \draw[black!55, line width=0.3pt] (-0.2,-0.28) -- (-0.13,-0.14) -- (-0.06,-0.28);
    \draw[black!55, line width=0.3pt] (0.06,-0.28) -- (0.13,-0.14) -- (0.2,-0.28);
    \draw[black!55, line width=0.3pt] (-0.13,-0.14) -- (0,0) -- (0.13,-0.14);
  }]
\node[box, text width=3.3cm, minimum height=1.4cm] (x) at (0.3,-0.8)
  {\textbf{Feature row $x$}\\[1pt] one job on one\\candidate hardware type\\[2pt]
   56 numeric and\\17 categorical features};
\node[box, minimum width=4.9cm, minimum height=1.85cm, fill=orange!8] (router) at (1.0,1.37) {};
\node[font=\scriptsize\bfseries, anchor=north] at ([yshift=-1pt]router.north) {Bin classifier $g$ \ \textmd{(gradient-boosted classifier)}};
\foreach \i in {0,1,2} \pic at ($(router.center)+(-1.15+0.5*\i,0.4)$) {tree};
\node at ($(router.center)+(0.55,0.27)$) {$\cdots$};
\pic at ($(router.center)+(1.15,0.4)$) {tree};
\foreach \h [count=\k] in {0.02,0.03,0.05,0.08,0.14,0.30,0.62,0.38,0.12,0.06,0.03,0.02,0.01} {
  \pgfmathsetmacro{\xx}{-1.3+0.2*(\k-1)}
  \ifnum\k=7 \def\c{orange!85!black}\else\def\c{orange!45}\fi
  \fill[\c] ($(router.south)+(\xx,0.4)$) rectangle ++(0.15,0.55*\h);
}
\draw[black!50] ($(router.south)+(-1.36,0.4)$) -- ($(router.south)+(1.36,0.4)$);
\node[anchor=north] at ($(router.south)+(0,0.38)$) {$P(\text{bin } k \mid x)$, \ $k = 1,\dots,13$};
\coordinate (sTL) at (4.6,0.65);  \coordinate (sBL) at (4.6,-0.35);
\coordinate (sTR) at (5.4,2.0);   \coordinate (sBR) at (5.4,-1.7);
\filldraw[fill=gray!15, draw=black!70, thick] (sTL) -- (sTR) -- (sBR) -- (sBL) -- cycle;
\node[align=center] at (5.0,0.15) {select\\bin};
\def\ex{7.35}
\foreach \k/\lab [count=\j] in {1/{$f_1$ \ (lowest)}, 2/{$f_2$}, 3/{}, 7/{$f_{\hat k}$}, 8/{}, 13/{$f_{13}$ \ (highest)}} {
  \pgfmathsetmacro{\yy}{1.75-0.64*(\j-1)}
  \ifnum\k=3 \node[minimum width=2.9cm, minimum height=0.4cm] (e\k) at (\ex,\yy) {$\vdots$};
  \else\ifnum\k=8 \node[minimum width=2.9cm, minimum height=0.4cm] (e\k) at (\ex,\yy) {$\vdots$};
  \else\ifnum\k=7
    \node[box, minimum width=2.9cm, minimum height=0.4cm, fill=blue!18, very thick, draw=blue!60!black, inner sep=1pt] (e\k) at (\ex,\yy) {};
  \else
    \node[box, minimum width=2.9cm, minimum height=0.4cm, fill=blue!5, inner sep=1pt] (e\k) at (\ex,\yy) {};
  \fi\fi\fi
  \node[anchor=west] at ([xshift=2pt]e\k.west) {\lab};
}
\foreach \k in {1,2,7,13} {
  \foreach \t in {0,1,2} \pic[scale=0.55] at ($(e\k.east)+(-0.75+0.25*\t,0.08)$) {tree};
}
\node[font=\scriptsize\bfseries, anchor=south] at ([yshift=1pt]e1.north) {13 experts \ \textmd{(gradient-boosted regressors)}};
\node[box, text width=2.0cm, minimum height=1.0cm] (out) at (10.75,0 |- e7) 
  {$\hat q = \exp\big(f_{\hat k}(x)\big)$\\[1pt] predicted throughput};
\draw[arr] (x.north) -- (x.north |- router.south);
\draw[arr] (x.east) -- ++(0.55,0) |- (4.6,0.0);
\path (sTL) -- (sTR) coordinate[pos=0.3125] (sIn);
\draw[arr] (router.east |- 0,1.8) -- node[above, pos=0.45]{$\hat k$} (sIn |- 0,1.8) -- (sIn);
\foreach \k in {1,2,13} \draw[->, black!40, dashed] (5.4,0 |- e\k.west) -- (e\k.west);
\draw[arr, blue!60!black] (5.4,0 |- e7.west) -- (e7.west);
\draw[arr] (e7.east) -- node[above]{$\log \hat q$} (out.west |- e7.east);
\end{tikzpicture}
}
\caption{The prediction model. For each candidate hardware type, the job's
feature row $x$ goes to the bin classifier, a gradient-boosted classifier that
outputs a probability for each of 13 throughput bins. The selector passes
$x$ to the expert of the most likely bin,
$\hat k = \arg\max_k g_k(x)$, a gradient-boosted regressor trained under
the pinball loss, which predicts log throughput. Exponentiating gives the
prediction (Equation~\ref{eq:serve}). Experts are ordered from the lowest
to the highest throughput range.}
\label{fig:cascade}
\end{figure}
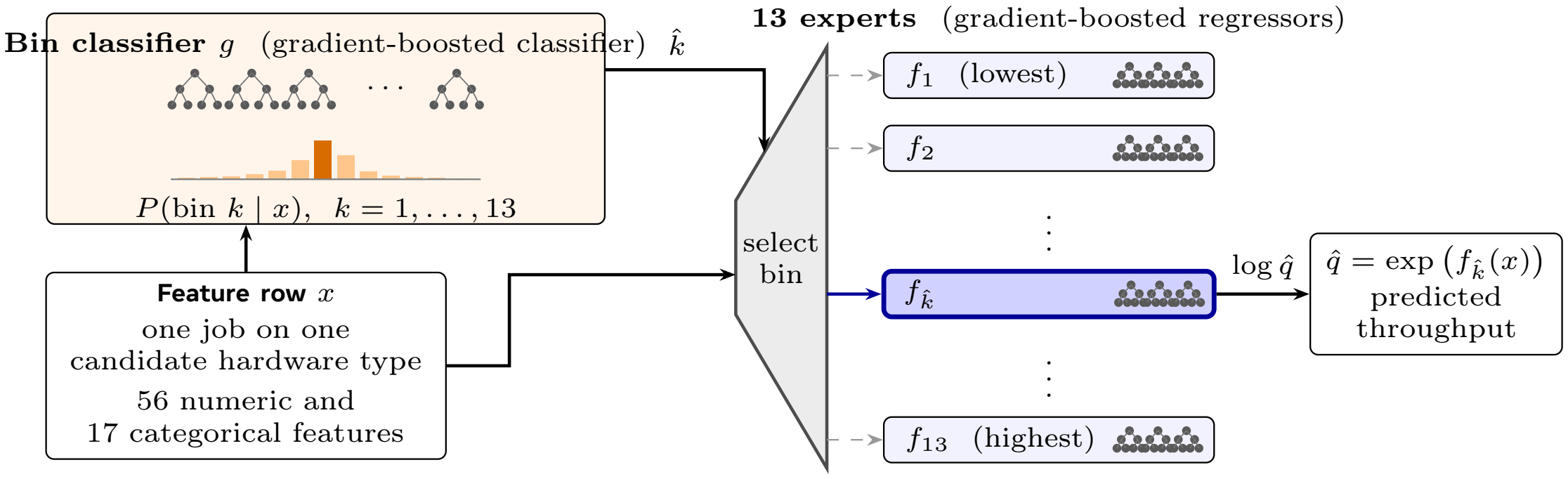

A single regressor over the full range still fits the extremes poorly, so
LEAP uses a binned cascade of gradient-boosted trees
(Figure~\ref{fig:cascade}). Training starts by splitting the log targets
into $K = 13$ equal-frequency bins. It then alternates two steps, as in
expectation maximization (EM). The first step fits one regressor, or
expert, per bin on that bin's rows:
\begin{equation}
f_k = \arg\min_{f} \sum_{i:\, z_i = k} \ell_{0.5}\big(y_i - f(x_i)\big),
\qquad k = 1,\dots,K,
\label{eq:mstep}
\end{equation}
where $z_i$ is the current bin of row $i$. The second step moves each row
to the bin whose expert predicts it best:
\begin{equation}
z_i \leftarrow \arg\min_{k} \big\lvert y_i - f_k(x_i) \big\rvert .
\label{eq:estep}
\end{equation}
The loop runs for at most three iterations and stops early when fewer than
2\% of rows change bin. The experts are then refit on the final bins, and a
gradient-boosted classifier $g$, the bin classifier, is trained to predict a row's
bin from its features alone, because $y$ is unknown at serving time. For a
job and candidate hardware type with feature row $x$, LEAP serves
\begin{equation}
\hat{q}(x) = \exp\big(f_{\hat{k}}(x)\big), \qquad
\hat{k} = \arg\max_{k}\, g_k(x),
\label{eq:serve}
\end{equation}
where $g_k(x)$ is the bin classifier's probability for bin $k$. Each training run
fits up to 52 regressors ($13 \times 3$ during EM and 13 in the final refit), and
the published model contains the 13 final experts and the bin classifier. The
cascade lets each expert specialize in one throughput range, while the
bin classifier only needs to place a job in the right range, which is an easier
problem than predicting its exact throughput.

The strongest features, the job's history, require special handling at
serving time. For each job family, LEAP uses the previous log throughput,
its running mean and standard deviation, the number of prior observations,
and the days since the last observation. Offline, these are a simple
group-by over the feature table. Online, they would require a keyed lookup
on the launch path. LEAP avoids this lookup by embedding, in each model
artifact, a snapshot of these statistics for up to 250,000 job families,
keyed on a normalized job name that strips run-specific identifiers. At
serving time only recency is recomputed, relative to the launch date. Jobs
whose family is absent from the snapshot are treated as cold starts and
receive an explicit imputed value rather than zero.

\subsection{Placement}
\label{sec:serving}

The prediction library is linked into the job launcher, so a routing call is
a function call on the launch path rather than a network request. For each job attempt, the launcher passes the job
configuration and its eligible hardware types. LEAP resolves the deployed model
pointer, loads the artifacts, builds one feature row per candidate by
adding that type's hardware specifications and the job's history features
from the snapshot, and predicts throughput on each candidate.

LEAP then scores each candidate by performance per unit cost, defined as
predicted throughput divided by the job's computational cost on that type. It removes premium hardware types unless
the job is predicted to use them efficiently, and returns the remaining types
in descending order of score. The cluster scheduler treats this list as a
preference rather than a constraint and resolves it against real-time
capacity. LEAP therefore decides which hardware type suits the job best,
while the final placement still depends on availability. Runtime switches
enable routing per model type, so routing can be rolled out, or rolled back,
one model type at a time.

\begin{figure}[t]
\centering
\includegraphics[width=0.55\linewidth]{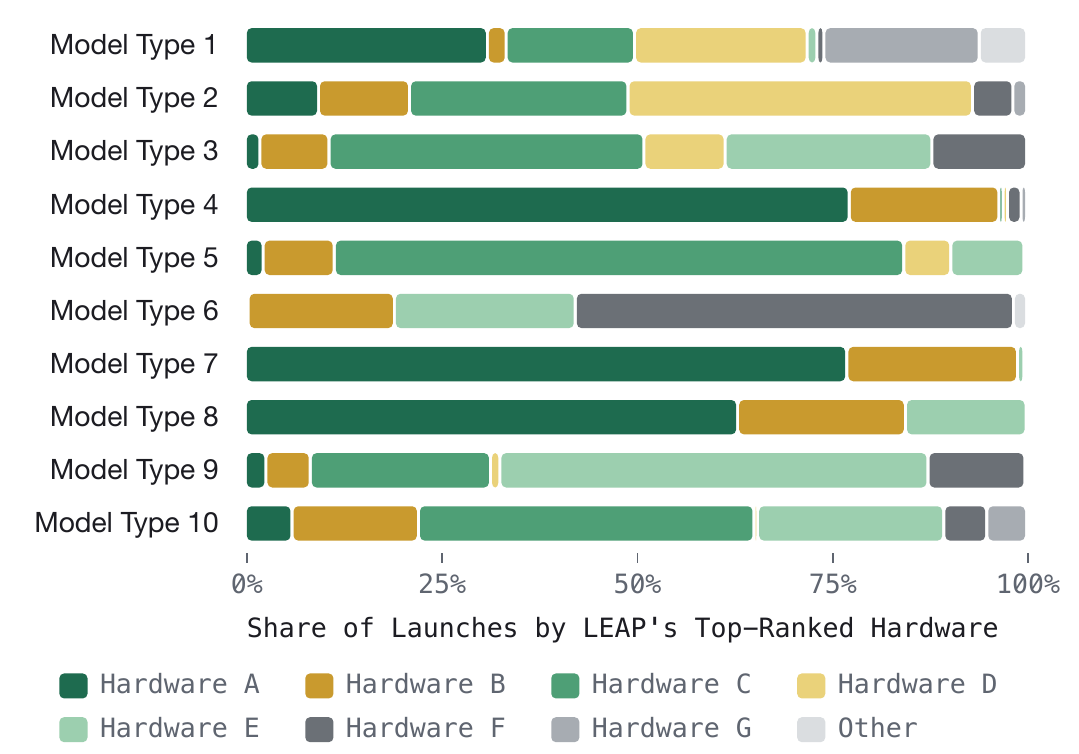}
\caption{Hardware type ranked first by LEAP, as a share of launches, for
ten model types over one week. Each launch makes one routing call; only
calls that returned a non-empty ranking are counted. A row with a single
color would indicate that one hardware type is always preferred for that
model type.}
\label{fig:place}
\end{figure}

Because the ranking is computed for each launch, launches of the same model
type do not necessarily receive the same hardware type.
Figure~\ref{fig:place} shows, for ten model types over one week, the share
of launches for which LEAP ranked each hardware type first. Every model
type has more than one top-ranked hardware type. For Model Type 1 (5,527
launches), four hardware types are each ranked first in 16--31\% of
launches. In four of the ten model types, no hardware type is ranked first
in more than half of the launches. In the most concentrated rows (Model
Types 4 and 7), the leading type accounts for 77\% of launches. Launches of
one model type differ in trainer count, batch size, prior throughput, and
the set of eligible hardware types, and these differences change which type
gives the highest throughput per unit cost. A static rule that maps each
model type to a single hardware type cannot express this variation.

Any failure, whether an empty candidate set, a missing feature, or an
artifact that cannot be loaded, degrades to an empty preference list. The
scheduler interprets an empty list as ``no opinion'' and places the job as
it would without LEAP.

\subsection{Nightly Retraining}
\label{sec:training}

A daily pipeline writes one partition of the feature table per day. Once a
partition arrives, a recurring job retrains the model for each prediction
target. Every job that LEAP places adds its measured throughput to the
table, so it becomes training data for the next night's model
(Figure~\ref{fig:LEAP_system}).

Each run trains on the last 30 days of data, about 750,000 job attempts, and holds out the most recent day for evaluation. All accuracy numbers in this
paper are measured on such held-out days, which the model did not see in
training. The window is computed relative to the run's scheduled date
instead of being fixed in the configuration. A fixed start date would
eventually fall outside the table's retention period and leave the run with
no training data.

Most of the training time goes to the cascade. A run fits 13 experts in
each of three EM iterations and once more on the final assignment, for 52
regressor fits, and then trains the bin classifier.

Each completed run writes an immutable, versioned artifact that contains
the experts, the bin classifier, and the history snapshot. The run then updates a
production pointer to the new artifact. Launchers load the model through
this pointer, so a new model reaches production without a code change. Test
runs do not update the pointer. Artifacts are kept for a limited time, so
an older result can be reproduced from its artifact only within that
retention period.

\section{Evaluation}
\label{sec:results}
\subsection{Experimental Setup}
\label{sec:protocol}

\paragraph{Training and Evaluation Data.} Each row of the dataset is one completed job attempt on
one hardware type. It contains the job's launch configuration, the
specifications of the hardware type, and the measured throughput, which is
the label (Section~\ref{sec:features}). Unless stated otherwise,
experiments follow the protocol that the deployed pipeline runs every
night (Section~\ref{sec:training}). A model is trained on the daily partitions of the previous 30 days, about 750,000 job attempts, and scored on the next day's partition, which it did not see in training. We do not remove outliers or clip targets; we drop only rows with missing or zero measured throughput, since the target is $\ln q$.

\paragraph{Metric.} Let $J$ be the set of all job attempts in the held-out
partition, across every model type and hardware type, and let $q_j > 0$ and
$\hat{q}_j$ be the measured and predicted throughput of attempt $j$.
Accuracy is reported as w20, the share of these attempts predicted within
$\pm 20\%$,
\begin{equation}
\mathrm{w20} = \frac{100}{|J|} \sum_{j \in J}
\mathbf{1}\!\left[ \frac{|\hat{q}_j - q_j|}{q_j} \le 0.2 \right].
\label{eq:w20}
\end{equation}
The evaluation pools the attempts of all model types in $J$, so w20 is not
measured on any single model type. If $J_m$ is the set of attempts of model
type $m$, w20 equals the attempt-weighted average of the per-model-type
accuracies,
\begin{equation}
\mathrm{w20} = \sum_{m} \frac{|J_m|}{|J|}\, \mathrm{w20}_m ,
\label{eq:w20-types}
\end{equation}
so model types that run more attempts contribute more. The weighting is by
attempt count, not by GPU-hours. Where we report tighter tolerances, w$k$
is defined in the same way with $0.2$ replaced by $k/100$.
We use w20 instead of MAPE or $R^2$ because both behave poorly on this
target. MAPE is a mean of ratios, so a few jobs with very small throughput
dominate it. $R^2$ on the raw scale is dominated by the largest jobs,
because the target spans several orders of magnitude, and on many nights it
is close to zero while w20 is above 75\%. Routing needs per-candidate
estimates that are close enough to rank hardware types correctly, and w20
measures that directly.
\paragraph{Repeated evaluation.} Because the model is retrained every
night, both the training window and the held-out jobs change from one night
to the next. We therefore report the headline accuracy over 30 consecutive
nightly retrains instead of a single run. Comparisons between methods score
every method on the same held-out day, since single-day differences of one
or two points are within the night-to-night variation.
\subsection{Results}
\label{sec:eval-results}
\subsubsection{Accuracy and Impact}
\label{sec:impact}
\begin{figure}[t]
\centering
\includegraphics[width=0.7\linewidth]{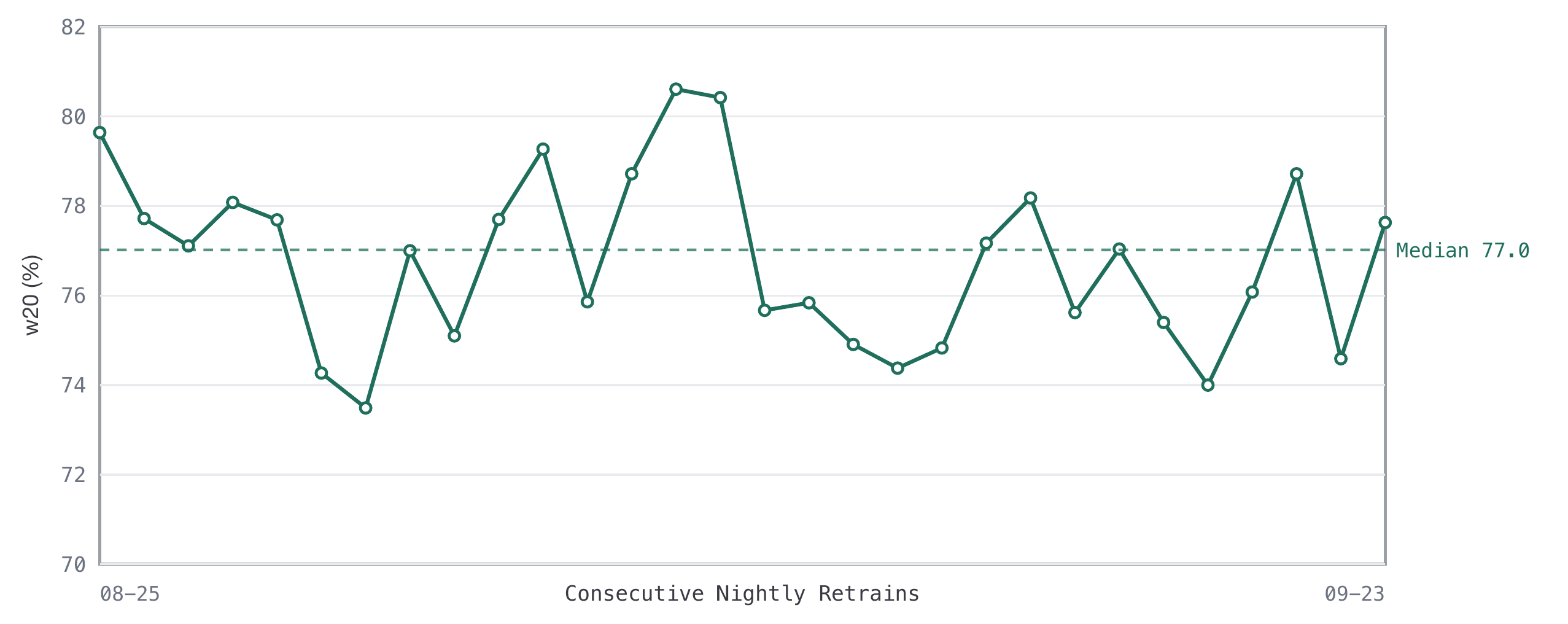}
\caption{w20 of 30 consecutive nightly retrains. Each model is trained on
the previous month and scored on the following day's partition. The median
is 77.0\%, and individual nights range from 73.5\% to 80.6\%.}
\label{fig:nightly}
\end{figure}

Figure~\ref{fig:nightly} shows w20 for each of the 30 nightly retrains. The
median is 77.0\%, which we use as the headline accuracy of LEAP, and
individual nights range from 73.5\% to 80.6\%. Accuracy is lower at tighter
tolerances. On the held-out day used in Section~\ref{sec:llm}, LEAP predicts
68.9\% of job attempts within $\pm 10\%$ and 42.9\% within $\pm 5\%$, and its
median absolute percentage error is 6.9\%.
Over the first half of
2026, the share of GPU-hours that received a ranking rose from 58\% to 95\%, which corresponds to 77 million
GPU-hours. Fewer jobs run on the top-ranked type because the scheduler treats the ranking as a preference and places a job on the top-ranked type only when
capacity allows (Section~\ref{sec:serving}). In a six-day window in August
2026 covering about 200,000 jobs, jobs that ran on the top-ranked type
accounted for 19.9\% of jobs, 30.0\% of GPU-hours, and 26.2\% of computational cost-weighted usage. The jobs that LEAP can route account for about half of
the computational training cost of the fleet it serves. Over the first half of 2026, we
estimate that routing with LEAP reduced the total computational infrastructure cost of
this fleet by about 3\%.
\subsubsection{Comparison with the Rule-Based Baseline}
\label{sec:baseline}
Before LEAP, throughput estimates came from a rule-based lookup table. We
scored both methods on the same held-out subset of one day's job attempts (7,131 jobs from 276 model
types). The rule engine returns an estimate for 946 jobs (13.3\%), and LEAP
returns one for every job. Most of the missing coverage comes from model
types with no entry in the rule table (93\% of the model types on that day).
On the 946 jobs that the rule engine covers, LEAP predicts 85.1\% within
$\pm 20\%$, compared with 53.2\% for the rule engine
(Table~\ref{tab:baseline}).
\begin{table}[t]
\caption{LEAP and the rule-based baseline on a held-out subset of one day (7,131 jobs,
276 model types). The rule engine returns an estimate for 946 jobs (13.3\%)
and LEAP for all of them; accuracy is compared on those 946 jobs. w$k$ is
the share of jobs predicted within $\pm k\%$ of measured throughput, and
ratio is the median of predicted over measured throughput.}
\label{tab:baseline}
\begin{center}
\begin{small}
\begin{tabular}{lrrrr}
\toprule
 & w10 & w15 & w20 & Ratio \\
\midrule
Rule engine & 33.4 & 45.0 & 53.2 & 1.11 \\
LEAP        & \textbf{72.1} & \textbf{80.6} & \textbf{85.1} & \textbf{1.00} \\
\bottomrule
\end{tabular}
\end{small}
\end{center}
\end{table}
The difference grows at tighter tolerances. Within $\pm 10\%$, LEAP is
correct on 72.1\% of these jobs and the rule engine on 33.4\%. LEAP is also
better calibrated on this subset, with a median predicted-to-actual ratio of
1.00, compared with 1.11 for the rule engine. These 946 jobs belong to the
model types that the table was built for, so LEAP is more accurate even on
the jobs that favor the table.
\section{Model Ablation Study}
\label{sec:ablation}
This section compares the baseline model with alternative features,
model families, and inputs. Within each comparison, all models use the same
training window and are scored on the same held-out day
(Section~\ref{sec:protocol}).
\subsection{Features and Model Structure}
\label{sec:levers}
All models in this subsection predict log throughput from the same feature
table and are trained on about one month of data that ends the day before
the evaluation day. The starting point is a scikit-learn gradient-boosted
tree regressor~\citep{pedregosa2011sklearn} trained with the pinball loss at
the median (Equation~\ref{eq:pinball}), which is the model LEAP used before
the changes described here. Its inputs are the job's launch configuration,
such as trainer count, batch size, and parallelism settings, and the
specifications of the candidate hardware type.
We then added three groups of features. Job-history features are the job's
previous log throughput, the running mean and standard deviation of its
earlier log throughputs, the number of earlier runs, and the days since the
last run (Section~\ref{sec:model}). Throughput-ratio features combine
configuration values that relate directly to throughput, such as the total
batch size across trainers and the ratio of data readers to trainers.
Architecture features are 44 summary values derived from the model
definition, such as layer counts and arithmetic intensity. We also changed
the model twice. The deeper histogram GBDT replaces the baseline regressor
with scikit-learn's histogram-based regressor, with up to 3,000 trees of up
to 255 leaves and a learning rate of 0.03. The binned cascade replaces the
single regressor with 13 such regressors and a bin classifier that assigns each job
to one of them (Section~\ref{sec:model}).
\begin{table}[t]
\caption{w20 of model and feature changes, scored on one fixed held-out day
in August 2026. The adopted changes were applied in order, each on top of the
row above it. Rows in the middle block were tried and not adopted. The bottom block compares the graph-based neural model
with the model on the same jobs over three other evaluation
days, averaged over ten seeds.}
\label{tab:levers}
\begin{center}
\begin{small}
\begin{tabular}{lr}
\toprule
Configuration & w20 (\%) \\
\midrule
GBDT baseline                        & 72.3 \\
$+$ job-history features             & 74.8 \\
$+$ throughput-ratio features        & 76.6 \\
$+$ architecture features            & 80.7 \\
$+$ deeper histogram GBDT            & 82.8 \\
$+$ binned cascade                   & \textbf{83.3} \\
\midrule
LightGBM                             & 69.4 \\
TabPFN v2 (GPU inference)            & 79.1 \\
Gemma 3 27B, fine-tuned              & 69.0 \\
Llama 4, LoRA fine-tuned             & 50.8 \\
\midrule
\multicolumn{2}{l}{\emph{Same jobs, three days, ten seeds}} \\
Baseline cascade model                   & 77.4 \\
DNNPerf, job features only           & 77.1 \\
DNNPerf, job features $+$ graph      & 77.4 \\
\bottomrule
\end{tabular}
\end{small}
\end{center}
\end{table}
Table~\ref{tab:levers} scores every configuration in its first two blocks on
the same held-out day. That day is easier than the average night in
Figure~\ref{fig:nightly}, so the values in the table are higher than the
77.0\% headline, and only differences within the table are meaningful. The
largest gain came from the architecture features, which added 4.1 points.
Job-history features added 2.5 points, and throughput-ratio features added
1.8 points. The deeper histogram GBDT added 2.1 points, and the binned
cascade brought the model accuracy to 83.3\%. In total, these changes
raised w20 by 11.0 points over the baseline.
\subsection{Alternative Model Families}
\label{sec:alt-models}
We also tried other model families on the same features (middle block of
Table~\ref{tab:levers}). LightGBM, another gradient-boosting library,
reached 69.4\%, below the scikit-learn baseline.
TabPFN v2~\citep{hollmann2023tabpfn,hollmann2025tabpfn} is a transformer pretrained on
synthetic tabular data that predicts from examples given in context instead
of being trained on the task. For each job, we gave it the most similar
training rows as context, using the baseline features. It reached 79.1\%,
which was below the deployed model, and it needs a GPU and a
retrieval step at serving time.
We also fine-tuned two generative language models to read a job as text and
output its throughput. Each job's launch configuration is pruned to the
fields that affect performance and summarized into at most 12,000 tokens.
Llama~4 Scout~\citep{meta2025llama4} was fine-tuned with LoRA on 80,000 jobs. Gemma~3 27B~\citep{gemma2025gemma3} was fine-tuned the same way with more aggressive
pruning. Both were much less accurate than the GBDT, at 50.8\% for Llama~4
and 69.0\% for Gemma~3. Section~\ref{sec:llm} tests language models as
encoders instead.
\paragraph{Graph-based neural model.}
We reimplemented DNNPerf~\citep{gao2023dnnperf}, which predicts performance
from the model's computation graph. Our version runs an attention graph
neural network over an operator graph derived from each job's launch
configuration, combines it with the job features encoded with piecewise
linear encoding~\citep{gorishniy2022embeddings}, and predicts log throughput
with a ResNet head trained with a Huber loss. Scored on the same jobs as the
baseline model, it reached 77.4\% w20. Without the graph it
reached 77.1\%, so the graph added 0.25 points (95\% confidence interval
$-0.21$ to $0.71$).
We did not deploy this model for three reasons. First, it needs a GPU at
serving time, while the GBDT runs on the CPU inside the launcher. For
tens of thousands of routing calls per day and no gain in accuracy, the GPU  cost is not justified. Second, it needs the computation graph of every job.
Training requires extracting the graph from each job's model definition and
storing a graph cache of about 400\,GB for our training window, and every
routing call must look up the job's graph. Third, it is slower. A routing
call must load the graph cache and run the network, while the GBDT predicts
from a feature vector it already has.
\subsection{Language-Model Encoders}
\label{sec:llm}
The fine-tuned generative models in Table~\ref{tab:levers} had to produce
throughput as text, and they were among the least accurate models we tested. A
language model can also be used as an encoder, which turns the job's
description into a vector that a small network maps to a prediction. We
tested this approach with Qwen3-Embedding~\citep{zhang2025qwen3embedding}.
Each job's features are written as text, for example ``num\_trainers: 64,
batch\_size: 512, \ldots'', and the encoder embeds this text into a
1,024-dimensional vector. An MLP head takes the vector together with the
standardized numeric features and predicts log throughput under a Huber
loss. We either kept the encoder frozen or fine-tuned it with
LoRA~\citep{hu2022lora} together with the head. All models in this
subsection, including the baseline cascade, were trained on the same
training window and scored on the same held-out day.
\begin{table}[t]
\caption{Text-encoder predictors and the baseline GBDT cascade, scored on
the same held-out day. Encoder rows use Qwen3-Embedding with 0.6B or 4B
parameters, either frozen or fine-tuned with LoRA. GBDT $+$ embedding adds
the encoder's embedding to the GBDT's input features. MdAPE is the median
absolute percentage error.}
\label{tab:llm}
\begin{center}
\begin{small}
\setlength{\tabcolsep}{4pt}
\begin{tabular}{lrrrr}
\toprule
Model & w5 & w10 & w20 & MdAPE \\
\midrule
GBDT cascade & \textbf{42.9} & \textbf{68.9} & \textbf{77.3} & \textbf{6.9} \\
\midrule
0.6B, frozen              & 30.4 & 52.4 & 68.1 & 11.5 \\
0.6B, LoRA                & 26.1 & 45.2 & 73.6 & 9.3 \\
0.6B, LoRA, large head    & 28.6 & 48.1 & 72.7 & 10.5 \\
4B, LoRA                  & 26.6 & 48.4 & 70.7 & 10.5 \\
\midrule
GBDT $+$ embedding        & 30.2 & 50.3 & 72.4 & 9.9 \\
\bottomrule
\end{tabular}
\end{small}
\end{center}
\end{table}
None of the encoder configurations reached the accuracy of the baseline
cascade model (Table~\ref{tab:llm}). With a frozen 0.6B encoder, w20 was 68.1\%.
Fine-tuning the encoder with LoRA raised it to 73.6\%, the most accurate
encoder result and 3.7 points below the cascade. A larger head did not help
(72.7\%), and a 4B encoder, about seven times larger, was less accurate
(70.7\%). The gap is larger at tighter tolerances. Within $\pm 5\%$, no
encoder configuration was correct on more than 30.4\% of job attempts,
compared with 42.9\% for the cascade, and the median error of the best
encoder was 9.3\%, compared with 6.9\%. A likely reason is that the encoder
reads numbers as text tokens, which do not preserve magnitude well, while
throughput depends mostly on numeric features such as trainer count and
batch size (Figure~\ref{fig:featimp}). Adding the encoder's embedding to the
GBDT's input features reached 72.4\% w20, 4.9 points below the baseline
model. The embedding therefore did not improve on the baseline model,
and using it would add an encoder call to every routing call.
\subsection{New Hardware and New Model Families}
\label{sec:unseen-hw}
We removed one hardware type or one model family from training and scored
the model on the held-out data (Table~\ref{tab:unseen}). Neither model
extrapolates to a hardware type prior to deployment. Holding out a
hardware type cost 37 points, while removing the same number of rows at
random cost under 2, even though the held-out specifications lie within the
training range. A new hardware type must therefore run deployed jobs
before LEAP ranks it, and the next nightly retrain then includes it
(Section~\ref{sec:training}). A new model family also lowers accuracy. On the one family where we
also refit the cascade, the cascade lost 49.7 points and the graph-based
model 37.0, likely because the cascade's bin classifier has no basis for assigning
an unfamiliar family to a bin (Section~\ref{sec:model}).
\begin{table}[t]
\caption{Change in w20 when data is held out of training. Graph-based rows
average three hardware types or families over three days and ten seeds; the
single-family rows compare both models on one family (312 attempts, one
day).}
\label{tab:unseen}
\begin{center}
\begin{small}
\setlength{\tabcolsep}{4pt}
\begin{tabular}{llr}
\toprule
Held out & Model & $\Delta$w20 \\
\midrule
3 hardware types       & Graph-based & $-37.0$ \\
Random rows, same size & Graph-based & $-1.7$  \\
3 model families       & Graph-based & $-22.2$ \\
\midrule
1 model family         & Cascade     & $-49.7$ \\
1 model family         & Graph-based & $-37.0$ \\
\bottomrule
\end{tabular}
\end{small}
\end{center}
\end{table}
The graph-based model is the only alternative that generalizes better than
the cascade, and only to new model families. LEAP already predicts well for
a new configuration of a known model, because the job-history and
model-type features carry over. The difficult case is a new model whose
computation graph differs from everything in training. A future version of
LEAP could detect that case and send only those jobs to a graph-based
model.
\subsection{Summary}
\label{sec:ablation-summary}
No alternative we tested was more accurate than the baseline cascade model on
the same jobs. LightGBM, TabPFN, the fine-tuned language models, and the
language-model encoders were all less accurate, and the graph-based model
matched it. The cascade is also the best to serve. It runs on the CPU
inside the launcher, needs no GPU, graph cache, retrieval step, or encoder
call, and predicts from features that are already available at submission.

\section{Related Work}
\label{sec:related}
\paragraph{Performance models.}
Analytical models derive runtime from a model graph and hardware
constants~\citep{qi2017paleo, williams2009roofline, isaev2023calculon}.
Learned models scale a measured profile across
GPUs~\citep{yu2021habitat}, apply graph neural networks to computation
graphs~\citep{gao2023dnnperf, kaufman2021tpu, wang2025scaledl}, or model
per-kernel cost~\citep{lin2022dlrmperf}. The most accurate predictors use
execution traces, microbenchmarks, or live
profiling~\citep{liang2025lumos, chen2026hase}, and host-side data loading
can also limit throughput~\citep{pourali2026orion}. Each of these methods
needs a profiling run, a trace, or the job's computation graph. LEAP has
only the launch configuration, the hardware specifications, and the job's
history. Our DNNPerf re-implementation matched LEAP's accuracy, and it generalized
poorly to new hardware types (Section~\ref{sec:ablation}). Compiler cost
models~\citep{chen2018learning, zheng2020ansor} rank candidate schedules for
one kernel on one device and do not predict the absolute throughput of a
whole job.
\paragraph{Scheduling and cluster management.}
Heterogeneity-aware schedulers~\citep{narayanan2020gavel, jayaram2023sia}
and elastic schedulers~\citep{qiao2021pollux, peng2018optimus,
bankar2025coreml} assume per-type throughput is known or measure it while
the job runs. Cluster managers place work across heterogeneous
fleets~\citep{verma2015borg} and use learned predictions of lifetime,
runtime, or job objectives~\citep{cortez2017resource, park20183sigma,
jyothi2016morpheus}. Configuration search chooses hardware for a workload
from trial runs~\citep{venkataraman2016ernest, alipourfard2017cherrypick,
yadwadkar2017paris}. LEAP addresses the step before scheduling. It
estimates throughput on every candidate before the job runs, without trial
runs, and passes the scheduler a preference list.
\paragraph{Tabular learning.}
Gradient-boosted trees~\citep{chen2016xgboost, ke2017lightgbm,
pedregosa2011sklearn} remain strong for tabular regression and often match
deep tabular models~\citep{gorishniy2021revisiting, grinsztajn2022tree},
especially when numeric features are encoded
well~\citep{gorishniy2022embeddings}. In-context
transformers~\citep{hollmann2023tabpfn,hollmann2025tabpfn} and LLM
embeddings~\citep{tang2024llmembed} are recent alternatives. On our data,
a tuned GBDT was more accurate than both (Section~\ref{sec:ablation}).
\paragraph{Learned systems in production.}
Learned components have replaced hand-tuned heuristics in query
optimizers~\citep{marcus2019neo, marcus2021bao} and datacenter
scheduling~\citep{ding2026acela}, and Bao showed that such a component can
reach production when the existing system keeps control. LEAP follows this
pattern: the cluster scheduler keeps the final placement decision.
Production ML systems also suffer from hidden coupling and training-serving
skew~\citep{sculley2015hidden, breck2019datavalidation, baylor2017tfx}.

\section{Discussion and Future Work}
\label{sec:lessons}

\paragraph{Evaluation that matches deployment.}
An earlier internal evaluation scored the deployed model on rows from inside
its own training window and overstated accuracy by more than 20 points
compared with scoring the same model on later days. Every accuracy number in
this paper is measured on data the model did not see in training, and the
nightly results are scored on the day after the training window. Accuracy also varies by about seven
points from night to night (Figure~\ref{fig:nightly}), so a single-day
comparison between two models can reverse on the next day. We now compare
model variants only on the same held-out day and treat differences of one or
two points as inconclusive.

\paragraph{Model choice.}
In our experiments, increasing the capacity of a gradient-boosted tree model
and adding job-history features improved accuracy more than changing the
model family (Table~\ref{tab:levers}). TabPFN, fine-tuned language models,
and language-model encoders were less accurate. A graph-based neural model
reached the same accuracy but needs a GPU at serving time, a cache of
computation graphs, and more time per routing call. For a predictor that
runs on every job launch, these serving requirements weighed as much as
accuracy in our decision, and the GBDT runs on the CPU inside the launcher
with no additional infrastructure.

\paragraph{Limitations.}
LEAP learns from the fleet's own history, so it does not rank a hardware
type reliably until that type has run actual jobs
(\ref{sec:unseen-hw}). Accuracy on model families absent from training is
also low. w20 weights every job attempt equally, so it reflects the many
small jobs more than the few large jobs that account for most GPU-hours. 
\paragraph{Future work.}
Four directions follow from these results. First, the serving path should
supply the parallelism and scaling arguments that the model uses, or the
model should stop using them, and the comparison between served and offline
features should run every day. Second, jobs from a model family whose
computation graph differs from everything in training could be detected at
submission and sent to the graph-based model, which degrades less on such
families (\ref{sec:unseen-hw}). Third, a new hardware type could be
bootstrapped by running a small set of representative jobs on it before it
enters routing. Fourth, measuring the label-noise ceiling for throughput
would show how much of the remaining error a better model can recover.

\section{Conclusion}
\label{sec:conclusion}

We described LEAP, which predicts the throughput of each training job
on each candidate hardware type at submission time and ranks the candidates
by throughput per unit cost for the cluster scheduler. LEAP replaced a
lookup table that returned an estimate for 13\% of job attempts. LEAP
returns an estimate for every attempt it receives, and on the attempts the
table covered it predicts 85.1\% within $\pm 20\%$, compared with 53.2\% for
the table. Over 30 nightly retrains, its median w20 is 77.0\%. During the
first half of 2026, the share of fungible training GPU-hours that received a
LEAP ranking grew from 58\% to 95\%, and we estimate that routing with
LEAP reduced computational training cost by about 3\%.

\clearpage
\newpage
\bibliographystyle{assets/plainnat}
\bibliography{paper}

\end{document}